\documentclass[letterpaper, 10 pt, conference]{ieeeconf}  % Comment this line out if you need a4paper

\usepackage{xcolor}
\usepackage[pdftex]{graphicx}
\IEEEoverridecommandlockouts                              % This command is only needed if 
\usepackage{graphics} % for pdf, bitmapped graphics files
\usepackage{epsfig} % for postscript graphics files
\usepackage{mathptmx} % assumes new font selection scheme installed
\usepackage{times} % assumes new font selection scheme installed
\usepackage{amsmath} % assumes amsmath package installed
\usepackage{amssymb}  % assumes amsmath package installed

\usepackage{comment}
\usepackage[hidelinks]{hyperref}

\title{\LARGE \bf
A Convex Formulation of the Soft-Capture Problem
}

\author{Ibrahima Sory Sow, Geordan Gutow, Howie Choset, Zachary Manchester% <-this % stops a space
\thanks{*Ibrahima Sory Sow was supported by a Fellowship of the Belgian American Educational Foundation.}% <-this % stops a space
\thanks{Authors are with the Robotics Institute and Department of Mechanical Engineering, Carnegie Mellon University, 5000 Forbes Ave., Pittsburgh, PA 15213. {\tt\footnotesize \{isow, ggutow, choset, zmanches\}@andrew.cmu.edu}}%
}

\usepackage{lipsum} 

\usepackage{optidef}
\usepackage{algorithmicx}
\usepackage{algorithm,algpseudocode}

\usepackage{booktabs} %for tables

\usepackage[]{mdframed}
\usepackage{gensymb}

\newcommand{\half}{\frac{1}{2}}
\newcommand{\R}{{\mathbb{R}}}

\newcommand{\skewmat}[1]{[#1]^\times}

\begin{document}

\maketitle
\thispagestyle{empty}
\pagestyle{empty}

%%%%%%%%%%%%%%%%%%%%%%%%%%%%%%%%%%%%%%%%%%%%%%%%%%%%%%%%%%%%%%%%%%%%%%%%%%%%%%%%
\begin{abstract}

We present a fast trajectory optimization algorithm for the soft capture of uncooperative tumbling space objects. Our algorithm generates safe, dynamically feasible, and minimum-fuel trajectories for a six-degree-of-freedom servicing spacecraft to achieve soft capture (near-zero relative velocity at contact) between predefined locations on the servicer spacecraft and target body. We solve a convex problem by enforcing a convex relaxation of the field-of-view constraint, followed by a sequential convex program correcting the trajectory for collision avoidance. The optimization problems can be solved with a standard second-order cone programming solver, making the algorithm both fast and practical for implementation in flight software. We demonstrate the performance and robustness of our algorithm in simulation over a range of object tumble rates up to 10$^\circ$/s. %An open-source implementation of the algorithm is available at \todo{IN PROGRESS}. 

\end{abstract}

%%%%%%%%%%%%%%%%%%%%%%%%%%%%%%%%%%%%%%%%%%%%%%%%%%%%%%%%%%%%%%%%%%%%%%%%%%%%%%%%
\section{Introduction}\label{Sec:intro}
Many envisioned autonomous on-orbit interactions, including active debris removal, refueling, and repair or upgrade missions, require an active \textit{chaser} spacecraft to rendezvous and make contact with a passive or uncooperative \textit{target} orbiting body \cite{fehse_automated_2003}. A key requirement is \emph{soft capture}: near-zero relative velocity at the time of contact. Rendezvous and docking operations have historically been limited to cooperative targets that communicate with the chaser and control their state to facilitate docking \cite{woffinden_navigating_2007}. Such operations also require significant human interaction. However, capturing uncooperative or tumbling targets is also of significant interest. In such scenarios, human involvement is impractical due to the fast sense-plan-act loop required \cite{nasa_servicing_study}. Therefore, the chaser needs to carry out agile maneuvers autonomously. 

\begin{figure}[t!]
    \centering
    \includegraphics[width=0.99\columnwidth]{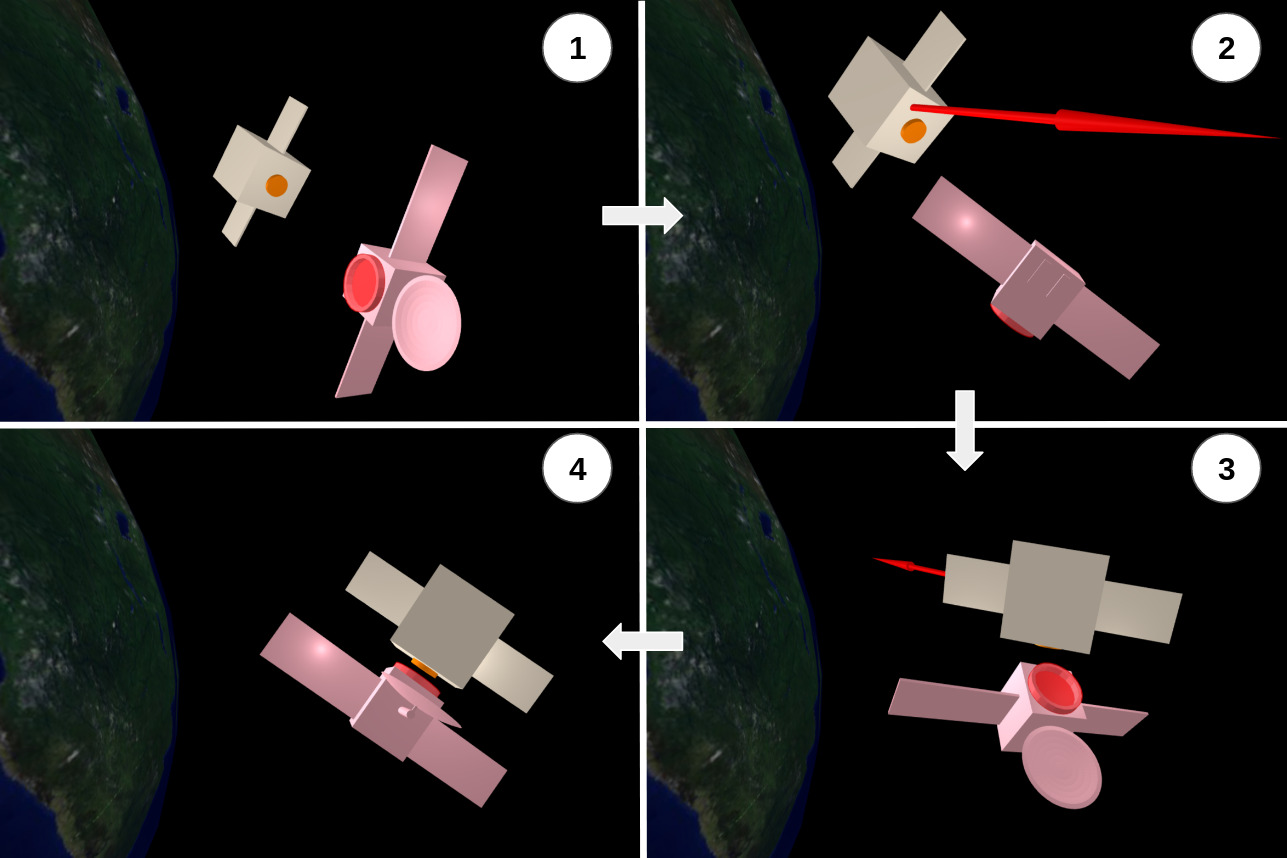}
    \caption{The soft-capture maneuver exhibits four phases: 1) The chaser (beige) approaches the target (pink) while maintaining its line of sight. 2) Impulsive burns help avoid collision with the client (thrust vector in red). 3) Corrective burns align the capture end-effector with the target capture frame. 4) the chaser is within the capture ball and turns on its capture end-effector.}
    \label{fig:first_page}
\end{figure}

A soft-capture mission would start with a stand-off phase estimating the dynamic properties and tumble state of the target. Next, during the approach phase (0-50m), the chaser needs to achieve the necessary relative states to initiate the capture phase, when the capture interface is activated to grapple the target. A significant body of work has focused on using a free-floating robotic manipulator to grapple a slowly rotating target with minimal base disturbance \cite{flores-abad_optimal_2017}. Virgili-Llop and Romano \cite{virgili-llop_simultaneous_2019} introduced a two-phase sequential-convex-programming (SCP) method for the capture of a tumbling target. For the approach phase, Sternberg and Miller \cite{sternberg_parameterization_2018} analyzed various parameterization schemes for creating fuel-efficient trajectories, whereas Stoneman and Lampariello \cite{stoneman_nonlinear_2016} devised a computationally expensive approach for generating offline reference trajectories for online tracking. Albee et al. \cite{albee_robust_2021} experimented with offline-generated trajectory look-up tables for online parsing and robust tracking. Malyuta et al. \cite{malyuta_fast_2019} proposed an SCP-based strategy for six-degree-of-freedom docking maneuvers against a stationary target. 

In this work, we focus on the approach phase, during which the majority of the mission $\Delta{V}$ is expended \cite{nasa_servicing_study}. Viewed as a trajectory optimization problem, inherent nonconvexities in the constraints must be addressed. For instance, field of view constraints (direct line of sight with the target) have historically been relaxed through semidefinite programming (SDP) methods \cite{kim_convex_2010} \cite{wu_relative_2010} or through Convex-Concave programming approaches \cite{shen_disciplined_2016}. For collision-avoidance constraints, relaxations have relied on pre-determined keep-out zones or rotating hyperplanes requiring precise prior timing knowledge \cite{reynolds2017small}.

Our approach leverages recent advances in convex optimization, which have been applied to many complex aerospace guidance and control problems, such as the classical problem of powered soft landing \cite{acikmese_lossless_2013}. Convex optimization offers deterministic guarantees given a desired solution accuracy \cite{boyd_bible} and has modest computational requirements, making it suitable for implementation onboard spacecraft.

We introduce a safe and efficient trajectory optimization algorithm for the chaser spacecraft to capture a target tumbling at up to 10$^\circ/\mathrm{s}$. Our algorithm produces minimum-fuel trajectories that respect stringent and nonconvex safety constraints: target within field-of-view, collision avoidance, and near-zero relative velocity at the time of capture. We first solve a convex problem by enforcing a convex relaxation of the field-of-view constraints, followed by a sequential convex program correcting the trajectory for collision avoidance. The optimizations can be solved with a standard second-order cone programming solver, making the algorithm both fast and practical for implementation in flight software. We evaluate the convergence behavior and performance of the algorithm on 250 different target tumble rates (up to 10$^\circ/\mathrm{s}$) and relative chaser orbits in simulation.

The paper proceeds as follows: Sec. \ref{Sec:background} reviews relevant mathematics. In Sec. \ref{Sec:rel-orbit} the problem setup and dynamics are detailed. Sec. \ref{Sec:formulation} presents the nonconvex problem and Sec. \ref{Sec:cvx} describes its convex relaxation. Sec. \ref{Sec:experiments} evaluates the computational, safety, and robustness performance of our algorithm, and Sec. \ref{Sec:conclusion} summarizes our conclusions.

\section{Background}\label{Sec:background}
%%%%%%%%%%%%%%%%%%%%%%%%%%%%%%%%%%%%%%%%%%%
\subsection{Quaternion Algebra}
We represent attitude with unit quaternions following the conventions in \cite{jackson_planning_2021}. We define a quaternion $q \in \R^4 := [q_s \;\; q_v^T]^T$ where $q_s \in \R$ and $q_v \in \R^3$ are its scalar and vector parts, respectively. We define the matrix
  \begin{align}
    L(q) := \begin{bmatrix} 
        q_s \;\; & -q_v^T \\ 
        q_v \;\; & q_s I + \skewmat{q_v} 
    \end{bmatrix} 
    \label{eq:Lmult}
\end{align} 
to compute the quaternion product $q_2 \otimes q_1 = L(q_2) q_1$. We use $Q(q)$ to denote the rotation matrix equivalent to $q$.

\subsection{Differentiable Collision Metric} \label{bg:dcol}
\begin{figure}[h!]
    \centering
    \includegraphics[width=0.99\columnwidth]{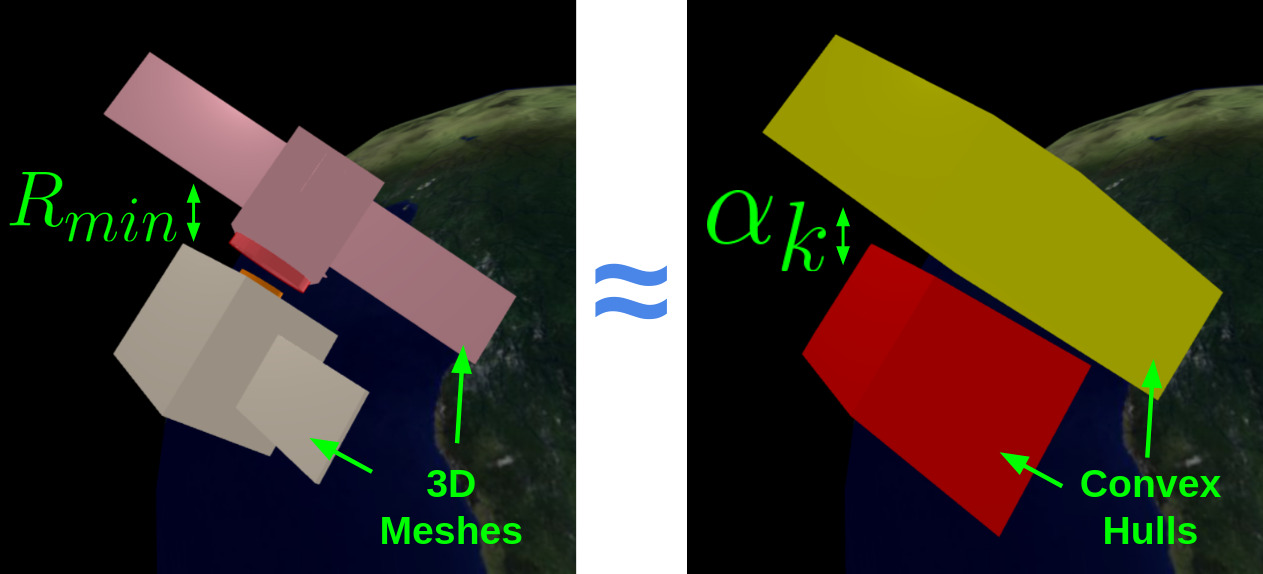}
    \caption{Left: Collision detection methods find a non-differentiable minimum distance between the 3D meshes. Right: DCOL \cite{tracy_differentiable_2023} finds the minimum inflating factor $\alpha$ between the convex hulls leading to an intersection with fast Jacobian computations.}
    \label{fig:dcol_illustration}
\end{figure}

Collision-avoidance constraints yield a nonconvex feasible region that is time-varying and dependent on both orbit and attitude. Traditional collision-detection routines are inherently non-differentiable, making them hard to incorporate in optimization routines. In this work, we conservatively approximate the chaser and target geometries by their convex hulls through a set of linear inequalities and adopt DCOL \cite{tracy_differentiable_2023}, a differentiable collision formulation visualized in Fig. \ref{fig:dcol_illustration}. DCOL solves a small convex optimization problem for the minimal inflation factor $\alpha$ leading to an intersection between the two inflated convex hulls. The approach is differentiable by design, and collision is detected if $\alpha < 1$. DCOL allows fast computation of the Jacobian of the inflation factor $\alpha$ with respect to the position and attitudes of both bodies \cite{tracy_differentiable_2023}, permitting its use for relaxations in sequential convex optimization.

\section{Relative Orbital Dynamics}\label{Sec:rel-orbit}
\begin{figure*}[!ht]
    \centering
    \includegraphics[width=0.95\linewidth]{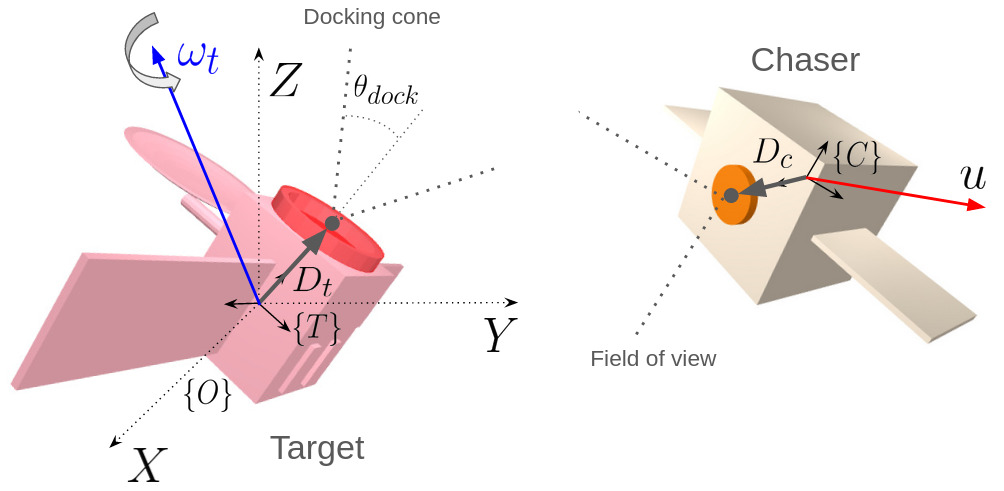}
    \caption{$\{O\}$ is the reference frame. The target body frame $\{T\}$, rotating with an angular velocity vector $\omega_t$, coincides with the center of $\{O\}$. $D_t$ is the capture point on the target vehicle. The chaser body frame $\{C\}$, with relative position $r$ and velocity $v$ expressed in $\{O\}$, is centered at the center of mass of the chaser. $u$ is the thrust vector. $D_c$ is the chaser capture point.}
    \label{fig:reference_frame}
\end{figure*}
We assume a passive (unactuated) and uncooperative target spacecraft in orbit around Earth. Its position coincides with the origin of a local-vertical-local-horizontal (LVLH) frame, or Hill frame, denoted $\{O\}$, where the X-axis points along the radial direction, the Z-axis along the orbital angular momentum vector, and the Y-axis completes the right-handed frame, pointing along the direction of motion. The attitude and angular rates of the target spacecraft make up the state $x_t(t) = [q_t\;\; \omega_t]$ and are described through a frame $\{T\}$ attached to its assumed center of mass as shown in Fig. \ref{fig:reference_frame}. The tumbling trajectory follows Euler's equations for rigid bodies and the quaternion kinematics equation:
\begin{equation} 
    x_t = \begin{bmatrix} q_t \\ \omega_t \end{bmatrix}, \quad 
    \dot{x_t} = \begin{bmatrix} 
        \half L(q_t) \begin{bmatrix} 0 \\ \omega_t \end{bmatrix}\\ 
        J^{-1}(- \omega_t \times J \omega_t) 
    \end{bmatrix} ,
    \label{eq:target_dyn}
\end{equation}
where $\omega_t$ is the angular velocity of $\{T\}$, $q_t$ the rotation from $\{T\}$ to $\{O\}$, and $J$ is the target's inertia matrix. 

We model the chaser spacecraft as a point mass with state $x(t) = [r(t), v(t)]^T = [r_x(t), r_y(t), r_z(t), v_x(t), v_y(t), v_z(t)]^T$ and a thrust vector $u(t) = [u_x, u_y, u_z]^T$. We decouple translation and attitude by assuming the presence of a low-level target-pointing attitude controller that can track any angular velocity reference $\omega(t)$ with magnitude less than $\omega_{max}$. Given the short duration of the maneuver compared to the orbital period, the chaser thrust represents the most significant perturbation, and other orbital perturbations can be safely neglected. Under the assumption of a circular orbit for the target, the chaser motion can be modeled with the Hill-Clohessy-Wiltshire equations \cite{clohessy_terminal_2012}
\begin{align}
\dot{v}_x &= 3n^2 r_x + 2n v_y + \frac{u_x}{m}  ,\nonumber\\
\dot{v}_y &= - 2 n v_x + \frac{u_y}{m} ,\\
\dot{v}_z &= - n^2 r_z + \frac{u_z}{m} ,\nonumber
\label{cw_dynamics}
\end{align}
which can be put in the standard matrix form
\begin{equation}
\dot{x}(t) = A x(t)+ B u(t) 
\label{cw_dynamics_matrix}
\end{equation}
where $m$ is the mass of the chaser spacecraft, $n = \sqrt{\mu/a^3}$ is the orbital mean motion of the target spacecraft, $\mu$ is the gravitational parameter for Earth, and $a$ is the semi-major axis of the target orbit. The continuous dynamics equations are integrated to yield the discrete-time dynamics
\begin{equation}
    x_{k+1} = A_d x_k + B_d u_k
    \label{eq:cw_dyn_discrete}
\end{equation}
We assume that the chaser spacecraft initiates its proximity maneuvers from a passively stable relative orbit \cite{nasa_servicing_study_2010}, i.e. a bounded relative orbit where the energies of the chaser and target orbits are matched. With a slightly different inclination and eccentricity from the target's orbit, maintaining this orbit requires minimal $\Delta$V. The set of initial conditions of the chaser is then restricted by
\begin{equation}
    \dot{r}_{y} = -2 n r_x, \quad r_y = \frac{2 \dot{r_x}}{n} ,\label{no_drift_and_offset} 
\end{equation}
respectively canceling out the relative linear drift and offset of the relative orbit. Using the closed-form solution of the Clohessy-Wiltshire equations \cite{clohessy_terminal_2012}, the set of states lie on a parameterized ``safe ellipsoid'' with
\begin{equation}
    A_0 = \sqrt{(\frac{r_x}{n})^2 + r_x^2}, \quad B_0 = \sqrt{(\frac{r_z}{n})^2 + r_z^2} \label{eq:B_0}
\end{equation}
where the radial, along-track, and cross-track dimensions of the ellipsoid are $A_0$, $2A_0$, and $B_0$, respectively. Combining \eqref{no_drift_and_offset} and \eqref{eq:B_0} allows us to sample passively stable relative orbits.

The chaser predicts the future nominal motion of the target by integrating Eqs. \eqref{eq:target_dyn} at a fixed timestep, where any state at $t > 0$ is accessed through cubic spline interpolation of the nominal trajectory.  While we have chosen a spacecraft as a target, our method extends to any other passively tumbling space object.

\section{The Soft-Capture Problem}\label{Sec:formulation}

The soft capture problem seeks state and control trajectories, $x(t)$ and $u(t)$, that drive the endpoint of the capture vector on the chaser spacecraft, $D_c$, to coincide with the endpoint of the capture vector on the target, $D_t$. The trajectories must minimize fuel consumption and satisfy mission-specific and safety constraints given the target state prediction $x_t(t)$. We pose this kinodynamic planning problem as a trajectory optimization problem of the form
\begin{mini}
  {x_{1:N}, u_{1:N-1}}{\ell_N (x_N) + \sum_k^{N-1} \ell_k (x_k, u_k)}{}{}
  \label{general_optimization}
  \addConstraint{x_{k+1}}{ = A_d x_k + B_d u_k}{}
  \addConstraint{f(x_k, u_k)}{=0}{}
  \addConstraint{g(x_k, u_k)}{\leq 0}{}
\end{mini}
where $x_k$ and $u_k$ are the decision variables, $\ell_k$ and $\ell_N$ are the stage and terminal costs, and $f$ and $g$ specify equality and inequality constraints. If the objective and the constraints are convex in $x_k, u_k$, the problem is a convex optimization problem, which can be solved to global optimality in polynomial time \cite{boyd_bible}.

To begin with, we upper bound the thrust and velocity magnitudes to reflect thruster limitations and operational safety constraints:
\begin{align}
    \lVert u_k \rVert_2 & \leq U_{max} \label{eq:c_umax} \\
    \lVert{} v_k\rVert{}_2 & \leq v_{max} \label{eq:c_vmax}
\end{align}

Standard docking maneuvers require the chaser vehicle to approach the target in a cone-shaped corridor originating from the docking port of a relatively static target \cite{fehse_automated_2003}. For our tumbling target scenario, we restrain the position of the chaser to be within a cone of half-angle $\theta_{dock}$ (see Fig. \ref{fig:reference_frame}) emanating from the capture point of the target spacecraft for the last $N_{dock}$ timesteps. The resulting second-order cone constraint is 
\begin{equation}
    \lVert W X_k \rVert_2  \leq \frac{c^T X_k}{\tan(\frac{\pi}{2}- \theta_{dock})} \label{eq:c_docking}
\end{equation}
where $X_k$ is the position of the chaser relative to the target capture point and $W$ and $c$ define the orientation and origin of the cone of the same target capture point
\begin{align}
        W &= R_k \begin{bmatrix}
        1 & 0 & 0 \\
        0 & 1 & 0 \\
        0 & 0 & 0   
        \end{bmatrix} \nonumber, \quad 
        c = R_k \begin{bmatrix} 0 \\ 0 \\ 1 \end{bmatrix} \nonumber 
\end{align}
where $R_k = Q(q_t(k \Delta t))$ is the attitude of the target at the timestep $k$.

The chaser's attitude controller must maintain the line of sight of its optical sensors within some angular tolerance \cite{fehse_automated_2003}. We ensure this by restricting the angle between line-of-sight vectors at successive timesteps to be less than $\omega_{max} \Delta t$:
\begin{equation}
    \frac{r_k^T r_{k+1}}{\lVert r_k \rVert \lVert r_{k+1} \rVert} \geq \cos (\omega_{max} \Delta t)
    \label{eq:c_ncx_fov}
\end{equation}
Despite $0\leq{}\omega_{max} \Delta t\leq{}\pi/2$ in our scenario, the constraint remains nonconvex. Note that this constraint, combined with Eq. \eqref{eq:c_docking}, ensures correct alignment of the chaser for capture. 

At the final time $t_f$, the endpoints of the capture vectors $D_c$ and $D_t$ must coincide (Eq. \eqref{eq:c_final_pos}) given the final predicted target's attitude $Q_t(N \Delta t)$. The attitude controller ensures that $D_c$ points at the target using the minimum-norm rotation increment at each timestep, so the attitude of the chaser is a position-dependent function, $q(r)$, or its rotation matrix equivalent, $Q(q(r))$. Thus, $Q(q(r_N))= Q_t(N \Delta t) Q_y(-180^\circ)$. To protect the integrity of the capture mechanism and the target, we bound the relative velocities between the target capture point and the chaser's center of mass (Eq. \eqref{eq:c_final_vel})
\begin{align}
     \lVert r_N - Q_t(N \Delta t) (D_c + D_t) \rVert_2 &\leq \epsilon_p \label{eq:c_final_pos} \\
     \lVert v_N - \omega_t((N-1) \Delta t) \times  D_t \rVert_2 &\leq \epsilon_v 
     \label{eq:c_final_vel}
\end{align}
with $\epsilon_p$ and $\epsilon_v$ respectively the terminal position and velocity tolerances, guaranteeing the inequalities achievement while softening the constraints for the optimization solver.

The chaser must not collide with the target spacecraft during the trajectory. We use the collision metric described in Sec. \ref{bg:dcol} and require $\forall k : \alpha_k > \alpha_{min} > 1$, ensuring a minimal separation distance between the convex hulls. The collision-avoidance constraint at timestep $k$ is
\begin{equation}
     \alpha (r_k, q_k(r_k), q_t(k \Delta t)) = \alpha (r_k, q_t(k \Delta t)) > \alpha_{min}
    \label{eq:c_ncvx_collision}
\end{equation}
To reduce fuel consumption, we minimize the sum of the L1 norms of the thrust vectors. The L1 norm of the control is proportional to fuel consumption and a better fuel minimizer than the (perhaps more common) L2 norm \cite{ross_how_2004}. In addition, the reaction control system of the chaser employs impulsive burns, where the thrust magnitude is close either to zero or to the maximum limit \cite{simon_l1}. The L1 norm encourages such ``bang-off-bang'' solutions. With this final addition, we fully define Problem 1:
\begin{mdframed}
\textit{Problem 1: Nonconvex Soft Capture Problem}
\begin{mini} 
  {x_k, u_k}{\sum_k^{N-1} \left \| u_k \right \|_1}{}{}
  \addConstraint{\eqref{eq:cw_dyn_discrete}, \eqref{eq:c_umax}, \eqref{eq:c_vmax}, \eqref{eq:c_docking}}{}{}
  \addConstraint{\eqref{eq:c_ncx_fov}, \eqref{eq:c_final_pos}, \eqref{eq:c_final_vel}, \eqref{eq:c_ncvx_collision}}{}{}
  \nonumber
\end{mini}
\end{mdframed}

Problem 1 is a nonconvex trajectory optimization problem where nonconvexities arise from the field-of-view and collision-avoidance constraints. Nonlinear programming solvers, such as IPOPT \cite{wachter_implementation_2006} and SNOPT \cite{gill_snopt_2002}, could potentially solve Problem 1 directly. However, such solvers lack time-complexity guarantees and converge to the nearest local minimum to some initial guess, complicating their real-time implementation. 
\begin{figure*}[!ht]
    \centering
    \includegraphics[width=0.95\linewidth]{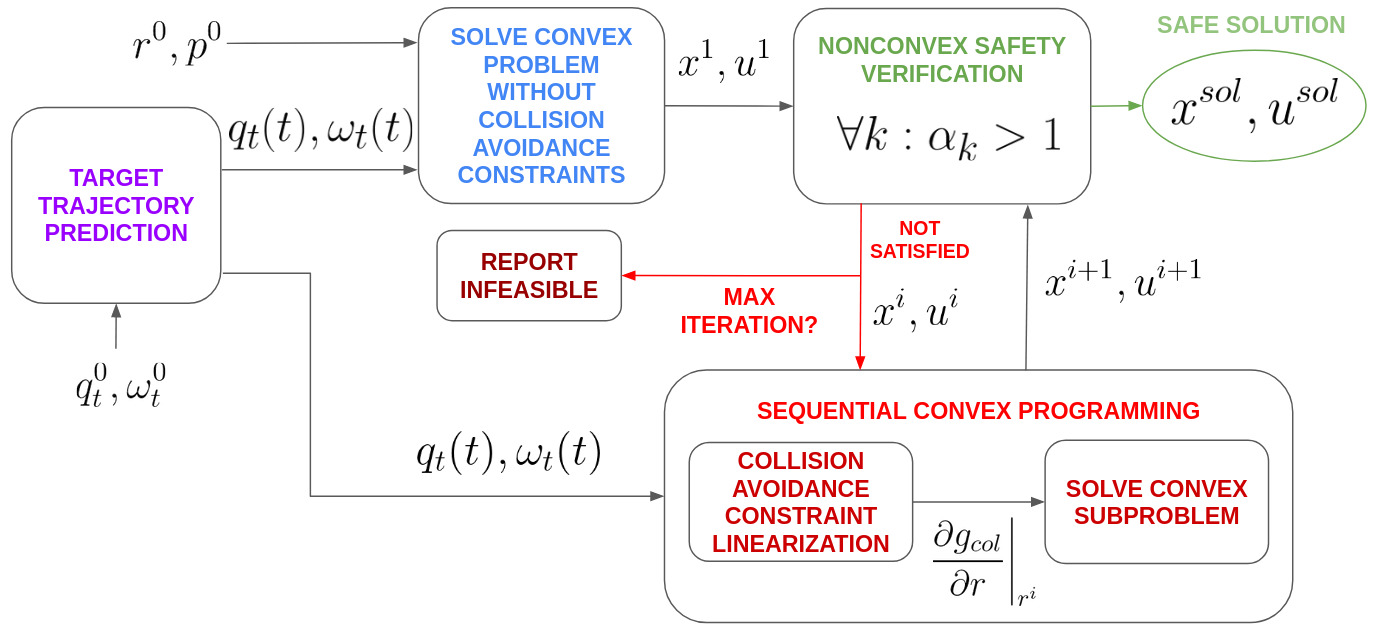}
    \caption{Illustration of the final algorithm. The future trajectory of the target $q_t(t), \omega_t(t)$ is predicted using the current estimate of the state of the target $q_t^0, \omega_t^0$. The prediction, along the current state of the chaser $r^0, v^0$, is input to an initial convex optimization problem without collision-avoidance constraints to obtain the first iterate of the trajectory $x^1, u^1$ for the sequential convex programming (SCP) pipeline. The SCP linearizes the collision constraints and iteratively solves convex subproblems until the safety criteria $\forall k: \alpha_k > 1$ is met to yield $x^{sol}, u^{sol}$. If the SCP reaches a set maximum number of iterations, the algorithm reports infeasibility.}
    \label{fig:scp_pipeline}
\end{figure*}

\section{Convex Trajectory Optimization}\label{Sec:cvx}

In this section, we convexify Problem 1 by relaxing the field-of-view \eqref{eq:c_ncx_fov} and collision-avoidance \eqref{eq:c_ncvx_collision} constraints.

\subsection{Field Of View} % \todo{capitalise all words in section titles}
We start by introducing an artificial bound constraint on the position vector $r_{max}$ and lifting the dimensions of the state vector $x$ by introducing the slack variable $\rho_k$ 
\begin{equation}
0 \leq \rho_k \leq r_{max}, \quad 
\left \| r_k \right \|_2 \leq \rho_k .\label{eq:c_rho}
\end{equation}
Replacing the norms in \eqref{eq:c_ncx_fov} by $\rho$ and multiplying by -1 gives
\begin{equation}
r_k^T r_{k+1} \leq  - \rho_k \rho_{k+1} \cos (\omega_{max} \Delta t) .\label{temp_in_exp}
\end{equation}
Minimizing $\rho_k$ in the objective guarantees a tight relaxation \cite{acikmese_lossless_2013}. We tackle the bilinearity by first noting that 
\begin{align}
\left \| r_k \right \|_2 \leq \rho_k &\implies{} r_k^T r_k \leq \rho_k^2 \label{sq1} ,\\
\left \| r_{k+1} \right \|_2 \leq \rho_{k+1} &\implies{} r_{k+1}^T r_{k+1} \leq \rho_{k+1}^2  .\label{sq2}
\end{align}
Adding the resulting terms (Eqs. \eqref{sq1}, \eqref{sq2}) respectively in both sides of Eq. \eqref{temp_in_exp} allows the expansion of a quadratic form
\begin{equation}
r_k^T r_k + r_k^T r_{k+1} + r_k^T r_{k+1} \leq \rho_k^2 - \rho_k \rho_{k+1} \cos (\phi) + \rho_{k+1}^2\label{temp_in_exp2} \nonumber ,
\end{equation}
where
$\phi = \omega_{max} \Delta t > 0$. This can be rewritten in matrix form 
\begin{equation}
z_k^T H z_k \leq y_k^T S y_k ,\label{temp_in_mtx}
\end{equation}
with
\begin{align}
z_k & = \begin{bmatrix}
r_k \\
r_{k+1} 
\end{bmatrix}, \quad
H = \begin{bmatrix}
I_{3x3} & 0.5 I_{3x3} \\
0.5 I_{3x3} & I_{3x3}
\end{bmatrix} > 0, \nonumber \\
y_k &= \begin{bmatrix}
\rho_k \\
\rho_{k+1} 
\end{bmatrix}, \quad
S = \begin{bmatrix}
1 & -0.5 \cos (\phi) \\
-0.5  \cos (\phi) & 1 \nonumber
\end{bmatrix} > 0.
\end{align}
As $H$ and $S$ are both symmetric and positive definite (since $\phi > 0$), we can simplify Eq. \eqref{temp_in_mtx}
%\Leftrightarrow  
\begin{equation}
\left \| \sqrt{H} z_k \right \|^2 \leq \left \| \sqrt{S} y_k \right \|^2 
\end{equation}
and by taking out the square
\begin{equation}
\left \| \sqrt{H} z_k \right \| \leq \left \| \sqrt{S} y_k \right \| 
\end{equation}
The eigenvalues of $S$ can be computed in closed-form and its smallest one is $\lambda_{min}(S) = 1 - cos(\phi)$. Subsequently, $\lambda_{min}(\sqrt{S}) = \sqrt{1 - cos(\phi)}$ and we can write 
\begin{equation}
    \left \| \sqrt{H} z_k \right \| \leq \lambda_{min}(\sqrt{S}) \left \| y_k \right \|_2 = \sqrt{1 - cos(\phi)} \left \| y_k \right \|_2 .
\end{equation}
Consecutive position vectors in time slightly differ in magnitude, $\rho_k \approx \rho_{k+1}$, and can be used to approximate $\left \| y_k \right \|_2$ 
\begin{equation}
     \left \| y_k \right \|_2 = \sqrt{\rho_k^2 + \rho_{k+1}^2} \approx \frac{1}{\sqrt{2}} (\rho_k + \rho_{k+1}) \nonumber ,
\end{equation}
which leads to
\begin{equation}
    \left \| \sqrt{H} z_k \right \| \leq \sqrt{\frac{1 - cos(\phi)}{2}} (\rho_k + \rho_{k+1}) \nonumber .
\end{equation}
Finally, with the trigonometric relation $sin^2(\phi) = \frac{1 - cos(\phi)}{2}$ and $1 - cos(\phi) > 0$, we derive the relaxed convex second-order cone field-of-view constraints:
\begin{equation}
    \left \| \sqrt{H} z_k \right \| \leq sin(\frac{\phi}{2}) (\rho_k + \rho_{k+1}) \label{eq:c_cvx_fov} .
\end{equation}

\subsection{Collision Avoidance}

We leverage the fast computation of Jacobians of the collision detection factor $\alpha_k$ (Sec. \ref{bg:dcol}) with respect to $r_k$ to linearize the nonconvex collision-avoidance constraints about a reference $r^i_k$, with the superscript $i$ denoting the iteration number. We further add a positive slack variable $s_k$, relaxing the inequality and ensuring its feasibility:
\begin{align}
    \alpha_k(r^{i}_k + \delta r_k) &\approx \alpha(r^{i}_k, q_t(k \Delta t)) + J_k^i \delta r_k + s_k > \alpha_{min} ,\label{eq:c_lin_dcol}
\end{align}
where %\frac{\partial \alpha}{\partial r}\bigg|_{r^{i}_k}
\begin{equation}
    J_k^i = \frac{\partial \alpha}{\partial r}\bigg|_{r^{i}_k} + \frac{\partial \alpha}{\partial q(r)}\bigg|_{r^{i}_k} \frac{\partial q(r)}{\partial r}\bigg|_{r^{i}_k} \nonumber ,
\end{equation}

We penalize $s_k$ in the objective function with a penalty weight $\psi$ as a measure of constraint infeasibility. For sufficiently large values of $\psi$, the objective term $\psi s_k$ defines an \textit{exact penalty function} \cite{NoceWrig06} and can be minimized by solving a sequence of subproblems.

\subsection{Final Algorithm}
%\todo{hyphenate compound modifiers}  

The final algorithm is depicted in Fig. \ref{fig:scp_pipeline}. With the chaser state and target trajectory prediction as input, we solve an initial convex second-order-cone program (SOCP) for $x, u$ with the convexified field of view constraint but without the collision-avoidance constraints (Problem 2). The near-feasible and near-optimal trajectory is then used to initialize a sequential convex subroutine solving for the corrections $\delta x, \delta u$ where the collision-avoidance constraint is added back and successively linearized along the previous solution (Eq. \eqref{eq:c_lin_dcol}), yielding Problem 3. We then update the current solution $i$
\begin{align}
    x_k^{i+1} = x_k^{i} + \delta x_k ,\quad 
    u_k^{i+1} = u_k^{i} + \delta u_k
\end{align}
This iterative convexification and correction scheme is repeated until a strictly safe trajectory $\forall k: \alpha_k>1$ has been achieved as illustrated in Fig. \ref{fig:scp_pipeline}. This choice of convergence criteria allows successful trajectories to go below the soft conservative limit $\alpha_{min}$.

The slack variable $s_k$ used in the collision-avoidance constraints ensures the feasibility of the subproblems and the exact penalty term in the objective function will succcessively eliminate any residual infeasibility. We report failure if Problem 2 is infeasible or if the SCP remains infeasible after the maximum allowable number of iterations (Problem 3), i.e. the solver has converged to a non-zero constraint violation. 

While not guaranteed to converge to a feasible solution, the algorithm is an efficient heuristic and can quickly detect an infeasible problem. The collision-avoidance constraints are applied in an uncrowded space where only the chaser and target are at risk of collision. Thus, as we show in Section \ref{Sec:experiments}, only small corrections are required from Problem 3 when the solution to Problem 2 is in collision. This permits the omission of a trust region. 

\begin{mdframed}
\textit{Problem 2: Convex Soft Capture Problem without collision-avoidance constraints}
\begin{mini}
  {x_k, u_k, \rho_k}{\sum_k^{N-1} \left \| u_k \right \|_1 + \sum_k^{N} \rho_k}{}{}
  \addConstraint{\eqref{eq:cw_dyn_discrete}, \eqref{eq:c_umax}, \eqref{eq:c_vmax}, \eqref{eq:c_docking},\eqref{eq:c_final_pos}}{}{}
  \addConstraint{\eqref{eq:c_final_vel}, \eqref{eq:c_rho}, \eqref{eq:c_cvx_fov}}{}{}
  %\label{eq:convex_problem}
  \nonumber
\end{mini}
\end{mdframed}

\begin{mdframed}
\textit{Problem 3: Convexified Soft Capture Sub-Problem with collision-avoidance constraints}
\begin{mini}
  {\delta x_k, \delta u_k, \rho_k, s_k}{\gamma \sum_k^{N-1} \left \| u^{i} + \delta u_k \right \|_1 + \sum_k^{N} (\rho_k + \psi s_k)}{}{}
  \addConstraint{\eqref{eq:cw_dyn_discrete}, \eqref{eq:c_umax}, \eqref{eq:c_vmax}, \eqref{eq:c_docking},\eqref{eq:c_final_pos}}{}{}
  \addConstraint{\eqref{eq:c_final_vel}, \eqref{eq:c_rho}, \eqref{eq:c_cvx_fov}, \eqref{eq:c_lin_dcol}}{}{}
  %\label{eq:convex_subproblem}
  \nonumber
\end{mini}
\end{mdframed}
\begin{figure}[t!]
    \centering
    \includegraphics[width=0.8\columnwidth]{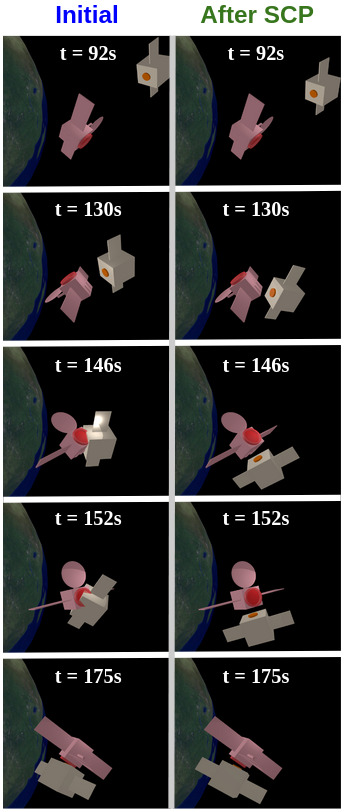}
    \caption{The initial trajectory (left) resulting from solving Problem 2 is infeasible. The SCP procedure (Problem 3) adjusts the ``tail'' of the trajectory to avoid collision with the target (right).}
    \label{fig:scp_traj_ba}
\end{figure}

\section{Numerical experiments}\label{Sec:experiments}

We evaluate our algorithm through two sets of experiments: an illustration of the corrective behavior of the SCP to satisfy the collision-avoidance constraint and an analysis of the computational and robustness properties of the algorithm for 250 pairs of sampled target initial tumble rates and chaser relative orbits. Numerical experiments were conducted on a laptop equipped with an Intel i7-1165G7 2.80GHz CPU and 16 GB of RAM using the optimization solver MOSEK \cite{mosek}, whose solve times are reported. All experiments were conducted using the Julia programming language. Parameters used for experiments are listed in Table \ref{table:sim_params}. The capture mechanism of the chaser is abstracted through the terminal position tolerance $\epsilon_p$. The target's inertia matrix is:
\begin{equation}
        J = \begin{bmatrix} 
        5.89056 & 0.0 & 0.0\\ 
        0.0 & 11.4462 & 0.233516 \\
        0.0 & 0.233516 & 11.5365 \\
        \end{bmatrix}
\end{equation}
 \begin{table}[ht!]
        \centering
        \caption{Simulation parameters}
         \begin{tabular}{l|l|l}
            \toprule[1pt]
            \addlinespace[1ex]
            Name & Symbol & Value \\
            \addlinespace[1ex]
            \midrule[1pt]
            \addlinespace[1ex]
            Semi-major axis of target's orbit & $a$ & 7.738e3 km\\ 
            Chaser mass & $m_c$ & 1500 kg \\
            Control cost penalty & $\gamma$ & 5\\
            Slack variable penalty & $\psi$ & 750\\
            Thrust bound & $U_{max}$ & 100 N \\
            Position bound & $r_{max}$ & 100 m \\
            Velocity bound & $v_{max}$ & 1.5 m/s \\
            Angular velocity bound & $\omega_{max}$ & 0.2 rad/s \\
            Terminal position tolerance & $\epsilon_p$  & 0.35 m \\
            Terminal velocity tolerance & $\epsilon_v$  & 0.03 m/s \\
            Number of docking constraints & $N_{fov}$ & 5 \\
            Docking cone half-angle & $\theta_{dock}$ & 30$^{\circ}$ \\
            Conservative collision factor & $\alpha_{min}$ & 1.3 \\
            Chaser capture vector in $\{C\}$ & $D_c$ & $\begin{bmatrix} 0 & 0 & 2.7 \end{bmatrix}$ m \\
            Target capture vector in $\{T\}$ & $D_t$ & $\begin{bmatrix} 0 & 0 & 2.7 \end{bmatrix}$ m \\
            Safety ellipsoid radial minimum & $A0_{min}$ & 15 m \\
            Safety ellipsoid radial maximum & $A0_{max}$ & 25 m \\
            Safety ellipsoid cross-track minimum & $B0_{min}$ & 10 m \\
            Safety ellipsoid cross-track maximum & $B0_{max}$ & 25 m \\
            Maximum iterations & $i_{max}$ & 15 \\
            Timestep & $\Delta t$ & 1s \\
            \addlinespace[1ex]
            \bottomrule[1pt] 
            \end{tabular}
        \label{table:sim_params}
\end{table}

\subsection{Sequential Convex Procedure Validation}

\begin{figure}
    \centering
    \includegraphics[width=\columnwidth]{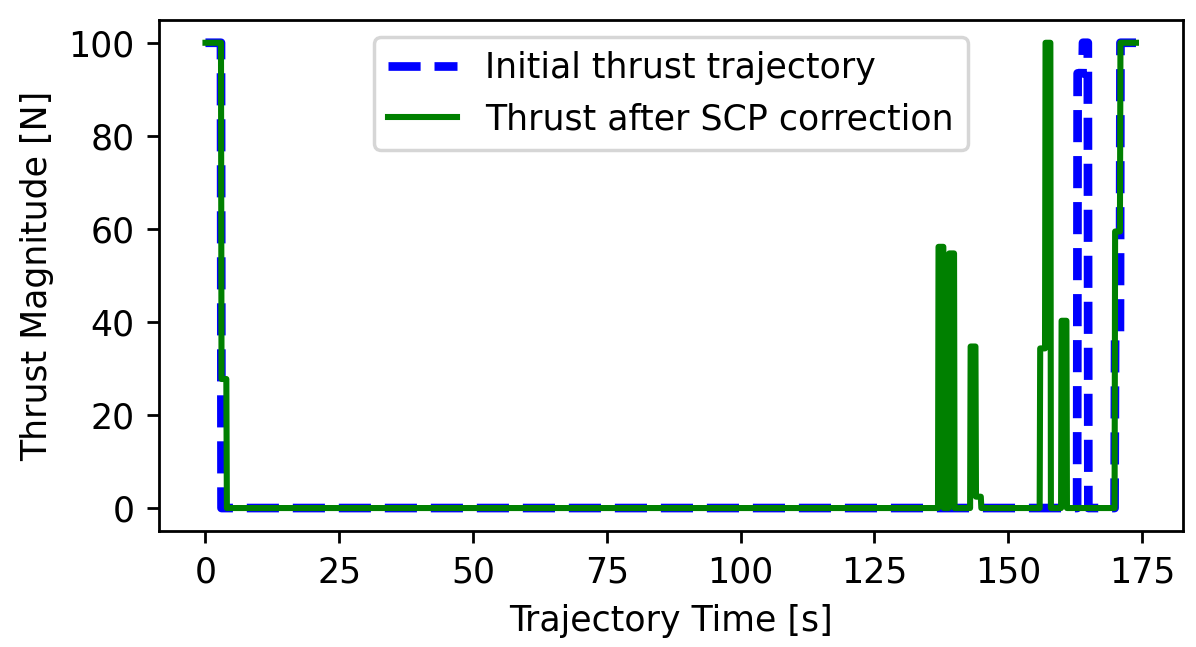}
    \caption{Nominal thrust trajectory without collision-avoidance constraints follows a distinct two-stage pattern: 1. a de-orbiting burn from the safety ellipse 2. final corrective maneuvers to align the chaser with the target. The sequential convex procedure adds additional $\Delta V$ maneuvers to the thrust profile to maintain safety.}
    \label{fig:scp_thrust}
\end{figure}

\begin{figure}
    \centering
    \includegraphics[width=\columnwidth]{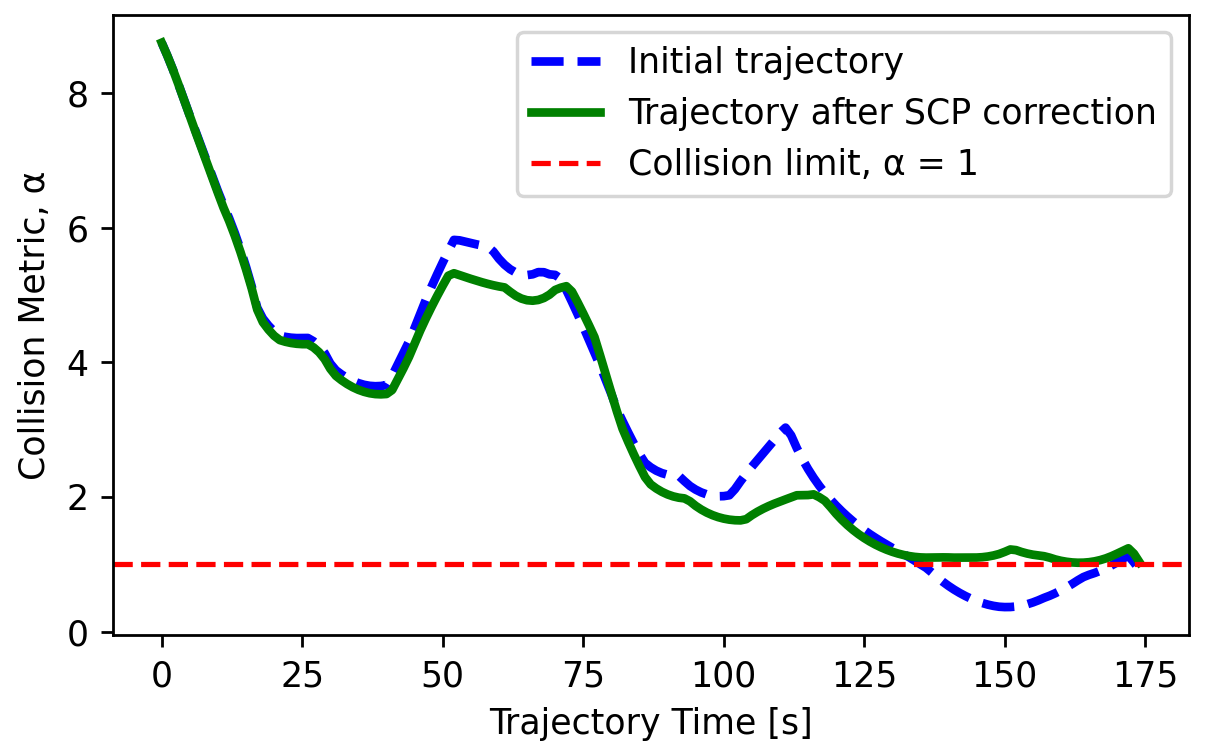}
    \caption{The initial convex solution is unsafe towards the end of the maneuver. The SCP corrects the trajectory into the feasible region.}
    \label{fig:scp_dcol}
\end{figure}

We illustrate the SCP corrective behavior on a single example (Fig. \ref{fig:scp_traj_ba}). The algorithm slightly adjusts the tail to make it feasible. We observed experimentally the nominal maneuver to require only two burns. Fig. \ref{fig:scp_thrust} displays the additional burns along the thrust profile required to avoid the large rotating keep-out regions defined by the convex hull of the target. Fig. \ref{fig:scp_dcol} shows the $\alpha$ histories; the end of the original trajectory is corrected above the collision threshold after four SCP iterations.
%\todo{need to make a stronger argument somewhere that this SCP idea will not fail.}

\subsection{Convergence and Robustness Analysis}

To evaluate the computational and robustness properties of our algorithm, we sampled 250 random attitudes and angular velocity vectors with initial tumble rates up to 10$^\circ/\mathrm{s}$ and paired them with 250 random relative orbits with a distance spanning from 15m to 50m. For each sample problem, we incremented the number of timesteps $N$ from $N_{min}$ to $N_{max}$ with a fixed timestep $\Delta t$ until the algorithm reached a solution. An infeasible problem was reported if no $N$ led to a successful trajectory.
\begin{figure}[h!]
    \centering
    \includegraphics[width=\columnwidth]{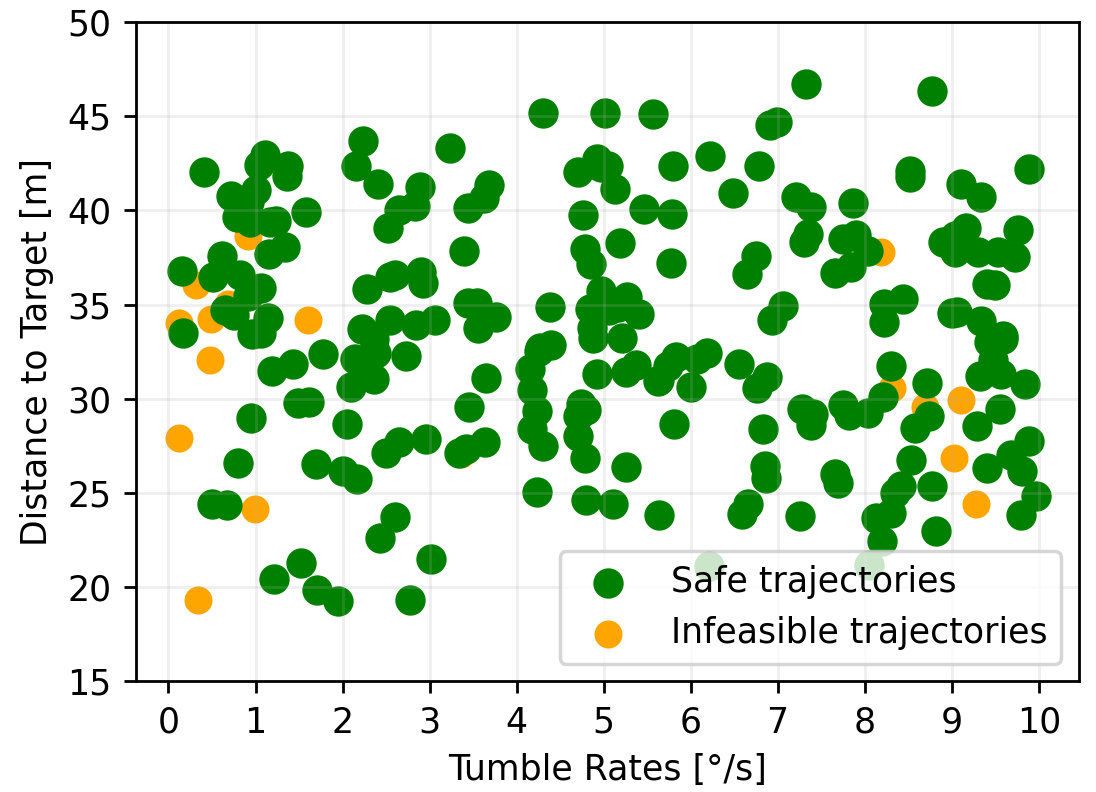}
    \caption{250 pairs of initial target tumble rate (up to 10$^\circ/s$) and attitude and chaser relative orbits (15m to 50m) were sampled as initial conditions. For 93.2\% (233) trajectories, the output of the algorithm resulted in a successful and safe capture of the target.}
    \label{fig:sampling_success}
\end{figure}  

\begin{figure}
    \centering
    \includegraphics[width=0.99\columnwidth]{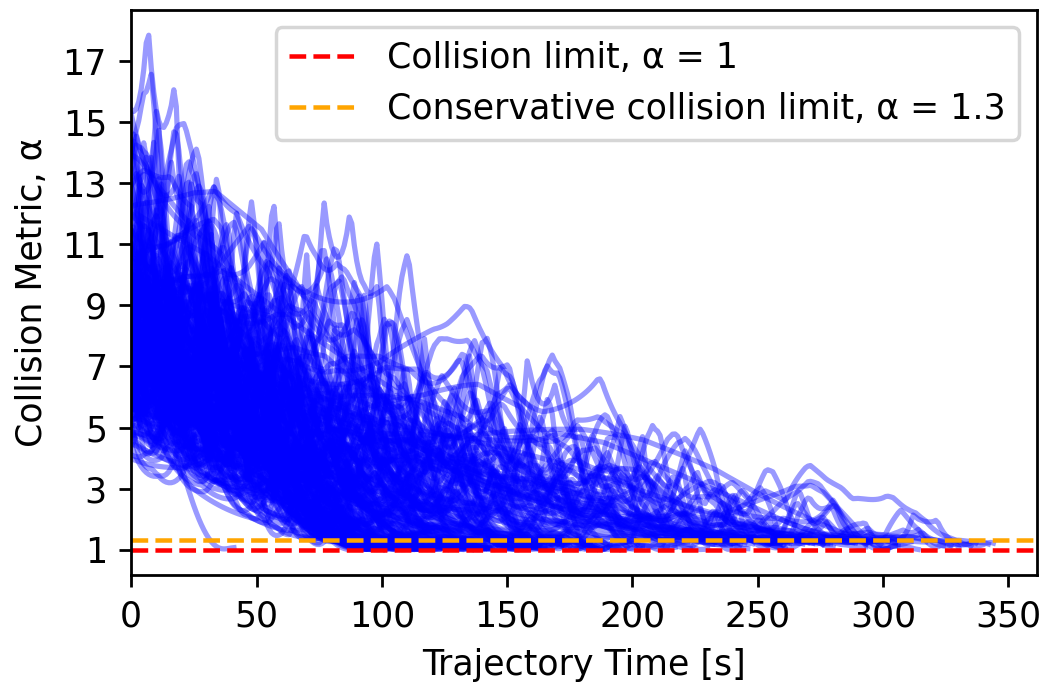}
    \caption{Depiction of all $\alpha(t)$ trajectories satisfying the hard collision constraint $\alpha_k > 1$ where $\alpha_{min} = 1.3$ defines a soft safety buffer zone.}
    \label{fig:dcol_traj}
\end{figure}

Fig. \ref{fig:sampling_success} shows that 93.2\% (233) of sampled initial conditions were successful while only 6.8\% (17) were infeasible within the bounds of the $N$-search ($N_{min}=40, N_{max}=350$). A per-iteration analysis of our algorithm showed an average runtime of 0.15 s per iteration. 8.6\% of successes resulted in safe trajectories after only Problem 2, not requiring any corrective actions from the SCP and ensuring global optimality (minimum-fuel). 54.9\% of sample trajectories required one SCP iteration for safety, while a cumulative 90.5\% of trajectories were safe after five SCP iterations. For a single SCP iteration, the average cumulative runtime was 0.336s, for five SCP iterations it was 0.889s. Much of the fast convergence can be attributed to the near-optimal initial guess after solving Problem 2. Fig. \ref{fig:dcol_traj} depicts the $\alpha_k$ evolution of all successful trajectories, confirming their 100\% safety guarantees.  We observe that using $\alpha_k$ as a proxy for distance, the trajectory time correlates with the distance which could be used to inform a final time $t_f$ search online. Note that our method applies in principle to tumble rates $\lVert{} \omega_t \rVert{} \leq \min(\omega_{max}, v_{max}/(D_c+D_t))$, as faster rates would require the chaser to violate its velocity bounds to achieve soft capture (0.2 rad/s using Table \ref{table:sim_params} values).
%A $\alpha_min = 1.3$ is used within the convex subproblem of the SCP to accelerate the convergence behavior of the algorithm.

\section{Discussion and Conclusions}\label{Sec:conclusion}

We introduced a fast algorithm to generate safe trajectories for the on-orbit soft capture of tumbling targets with operational and safety constraints. We convexified the field-of-view constraint and employed a sequential convex programming solver to handle the time-varying collision-avoidance constraints. We demonstrated the algorithm's computational and robustness properties for tumble rates up to 10$^\circ{}/\mathrm{s}$, showcasing its fast convergence and suitability for flight software implementation. Although we decoupled the translational and attitude dynamics, our algorithm still provides feasible reference attitudes for a low-level attitude controller (as a position-dependant function). Furthermore, it quickly detects if the initial conditions are infeasible.

We observed that the infeasible cases tended to occur at both very low and very high angular rates. On the one hand, we were able to render some of these cases feasible by adjusting the discretization parameters ($\Delta t$ and $N$). This suggests a need for a free-final time formulation whose naive implementation is, unfortunately, nonconvex. On the other hand,  the initial solution obtained by solving Problem 2 dictates the homotopy class of solutions accessible to the SCP part of our algorithm. Given the inherent local nature of the SCP procedure, we suspect that the optimal solution, particularly in low angular velocity regimes, resides in a different homotopy class and remains unreachable with our current method. 

Looking ahead, future work includes further investigation of the above limitations, i.e. developing a free-final-time formulation to expand the set of feasible solutions and addressing the homotopy class restriction imposed by our algorithm. On the safety side, plume impingement remains a concern. The plume effects define keep-out zones largely dependent on the thruster configuration used to generate the reference accelerations. Due to its similarity to the collision-avoidance constraints, a similar treatment as in this paper could be envisioned. Finally, uncertainties in the target's estimated state and dynamics could be accounted for in a variety of ways. An open-source implementation of the algorithm, along with examples, is available at: \url{https://github.com/RoboticExplorationLab/SoftCapture}

%This leads to time-varying target predictions, breaking the recursive feasibility of model-predictive control and posing significant challenges for potential on-board implementation.

%\addtolength{\textheight}{-12cm}   % This command serves to balance the column lengths
                                  % on the last page of the document manually. It shortens
                                  % the textheight of the last page by a suitable amount.
                                  % This command does not take effect until the next page
                                  % so it should come on the page before the last. Make
                                  % sure that you do not shorten the textheight too much.

%%%%%%%%%%%%%%%%%%%%%%%%%%%%%%%%%%%%%%%%%%%%%%%%%%%%%%%%%%%%%%%%%%%%%%%%%%%%%%%%

%%%%%%%%%%%%%%%%%%%%%%%%%%%%%%%%%%%%%%%%%%%%%%%%%%%%%%%%%%%%%%%%%%%%%%%%%%%%%%%%

%%%%%%%%%%%%%%%%%%%%%%%%%%%%%%%%%%%%%%%%%%%%%%%%%%%%%%%%%%%%%%%%%%%%%%%%%%%%%%%%
%\addtolength{\textheight}{-1cm}
\bibliographystyle{IEEEtran}

\bibliography{refs}

\end{document}